\documentclass[pdflatex,sn-mathphys-num]{sn-jnl}

\usepackage{graphicx}%
\usepackage{multirow}%
\usepackage{amsmath,amssymb,amsfonts}%
\usepackage{amsthm}%
\usepackage{mathrsfs}%
\usepackage[title]{appendix}%
\usepackage{xcolor}%
\usepackage{textcomp}%
\usepackage{manyfoot}%
\usepackage{booktabs}%
\usepackage{algorithm}%
\usepackage{algorithmicx}%
\usepackage{algpseudocode}%
\usepackage{listings}%
\usepackage{booktabs}
\usepackage{longtable}

\theoremstyle{thmstyleone}%
\theoremstyle{thmstyletwo}%

\theoremstyle{thmstylethree}%

\begin{document}

\title[Article Title]{StructDamage:A Large Scale Unified Crack and Surface Defect Dataset for Robust Structural Damage Detection}







\author[1]{\fnm{Misbah} 
\sur{Ijaz}}\email{25016119-001@uog.edu.pk}

\author*[2,5]{\fnm{Saif Ur Rehman} 
\sur{Khan}}\email{xuz72wot@rptu.de}
\author[1]{\fnm{Abd Ur} \sur{Rehman}}\email{a.rehman@uog.edu.pk}

\author[3,4,5]{\fnm{Sebastian} \sur{ Vollmer}}\email{sebastian.vollmer@dfki.de}
\author[2,3,4,5]{\fnm{Andreas} \sur{Dengel}}\email{andreas.dengel@dfki.de}

\author*[3,4,5,6]{\fnm{Muhammad Nabeel} \sur{Asim}}\email{muhammad\_nabeel.asim@dfki.de}

\affil[1]{\orgdiv{Department of Computer Science}, \orgname{University of Gujrat}, {\city{Gujrat}, \postcode{51700}, \country{Pakistan}}}

\affil[2]{\orgdiv{Department of Computer Science}, \orgname{Rhineland-Palatinate Technical University of
Kaiserslautern-Landau}, orgaddress{\city{Kaiserslautern}, \postcode{67663}, \country{Germany}}}

\affil[3]{\orgname{German Research Center for Artificial Intelligence}, \orgaddress{\city{Kaiserslautern}, \postcode{67663}, \country{Germany}}}

\affil[4]{\orgname{Intelligentx GmbH (intelligentx.com)}, \orgaddress{\city{Kaiserslautern}, \country{Germany}}}

\affil[5]{\orgname{BiogentX (biogentx.com}, \orgaddress{\city{Renala Khurd, District Okara, Punjab}, \country{Pakistan}}}

\affil[6]{\orgname{Department of Core Informatics, Graduate School of Informatics ,Osaka Metropolitan University}, \orgaddress{\city{Saka, 599-8531}, \country{Japan}}}

\abstract{Automated detection and classification of structural cracks and surface defects is a critical challenge in civil engineering, infrastructure maintenance, and heritage preservation. Recent advances in Computer Vision (CV) and Deep Learning (DL) have significantly improved automatic crack detection. However, these methods rely heavily on large, diverse, and carefully curated datasets that include various crack types across different surface materials. Many existing public crack datasets lack geographic diversity, surface types, scale, and labeling consistency, making it challenging for trained algorithms to generalize effectively in real world conditions. We provide a novel dataset, StructDamage, a curated collection of approximately 78,093 images spanning nine surface types: walls, tile, stone, road, pavement, deck, concrete, and brick. The dataset was constructed by systematically aggregating, harmonizing, and reannotating images from 32 publicly available datasets covering concrete structures, asphalt pavements, masonry walls, bridges, and historic buildings. All images are organized in a folder level classification hierarchy suitable for training Convolutional Neural Networks (CNNs) and Vision Transformers. To highlight the practical value of the dataset, we present baseline classification results using fifteen DL architectures from six model families, with twelve achieving macro F1-scores over 0.96. The best performing model DenseNet201 achieves 98.62\% accuracy. The proposed dataset provides a comprehensive and versatile resource suitable for classification tasks. With thorough documentation and a standard structure, it is designed to promote reproducible research and support the development and fair evaluation of robust crack damage detection approaches.}


\maketitle

\section{Background \& summary}\label{sec1}

Civil infrastructure, such as bridges, buildings, pavements, dams, and tunnels, forms the backbone of modern society because it provides trade, transportation, and key public services. Despite strict guidelines for construction and design, these structures ultimately suffer from deterioration due to factors such as severe weather, material wear and tear, environmental aging, and increased traffic demands \cite{chang2003health, alokita2019recent}. Their long term sustainability, dependability, and safety are seriously threatened by this degradation. Structural Health Monitoring (SHM) has become a vital and more significant approach to infrastructure maintenance and safety issues. Early detection of damage or deterioration allows for repair or replacement before more significant damage occurs. 
\cite{fujino2019research, martin2022facilitating}.

Cracks are an early indicator of structural damage. These parameters directly assess the structural health of infrastructure such as roads, buildings, bridges, pavements, and tunnels. Early maintenance and repair of cracks is crucial to prevent significant damage to infrastructure and surrounding areas. Manual crack inspection involves sketching crack patterns and identifying abnormalities \cite{bianchini2010interrater}. Manual crack detection relies on professional knowledge and experience, making it time consuming, costly, subjective, and unreliable. These limitations have driven increasing interest in automated crack detection approaches \cite{zeeshan2021structural}.

In recent years, Computer Vision (CV) based techniques have emerged as an efficient and non-destructive alternative for automatic crack damage detection. The rapid advancement of Machine Learning (ML) and Deep Learning (DL) techniques has resulted in considerable gains in crack detection, localization, and segmentation performance \cite{hamishebahar2022comprehensive}. However, the effectiveness of DL based models is dependent on the availability of high quality, diverse, and well annotated datasets, which is frequently sparse and difficult to get \cite{aung2025enhancing}. To create effective DL models for crack identification and segmentation, large datasets with various structural damage samples are needed.

Existing public datasets have individually contributed valuable data but are limited in scope. The METU concrete crack dataset \cite{METU} provides 40,000 images but is restricted to binary concrete crack classification. The road damage detection datasets (RDD2018-2022) \cite{RDD2018, RDD2019, Rdd2020, Rdd2022} offer large scale road inspection data but focus exclusively on pavement distress. The dacl10k dataset \cite{dacl10k} addresses bridge inspection but covers a different taxonomy of defects. No single resource has previously unified these complementary data sources into a cohesive multiclass benchmark encompassing the full spectrum of structural damage types encountered in practice.

StructDamage dataset addresses this gap by aggregating images from 32 publicly available datasets, spanning multiple surface types (concrete, asphalt, masonry), structural systems (pavements, bridges, buildings, dams), and geographic regions. Through careful curation, deduplication, quality filtering, and reannotation, we have produced a unified StructDamage dataset of approximately 78093 images organized into 9 surface damage categories. This dataset enables the development of robust, generalizable models for automated structural inspection and serves as a standardized benchmark for comparative evaluation of novel DL architectures.

\subsection*{Problem statement}

Although DL has significantly advanced crack detection, the field still lacks a unified benchmark that reflects the full range of damage types encountered in real world inspections. Most existing datasets are narrowly focused on a single material, structure type, or annotation task, which leads to fragmented research efforts and limited model generalization. As a result, models trained on one domain often perform poorly in others, and inconsistent labeling schemes prevent fair comparison across studies. In addition, several important damage categories such as spalling, corrosion, efflorescence, and vegetation growth, are missing from widely used benchmarks. In this work, we address these key challenges by integrating multiple public datasets into a single, harmonized framework. By harmonizing annotations, increasing data diversity, and applying systematic quality control, we deliver a comprehensive, benchmark ready dataset that supports fair comparison and more generalizable crack detection systems.

\subsection*{Research objective}
This dataset aims to serve as a standard resource to support robust, generalizable, and reproducible research in automated crack detection for SHM. To achieve this goal, this research focuses on the following objectives:
\begin{itemize}
    \item Create a comprehensive and unified novel dataset by integrating multiple public sources and standardizing heterogeneous annotation formats under a consistent multiclass taxonomy.
    \item Include diverse surface types, structural contexts, and previously underrepresented damage categories to improve model generalization.
    \item Apply systematic preprocessing, quality control, and transparent documentation to ensure dataset reliability.
    \item Provide a standard resource that supports fair evaluation, reproducible research, and development of robust, generalizable inspection models.
\end{itemize}
\subsection*{Contributions}
This work introduces novel StructDamage dataset, a largescale, multiclass structural damage dataset, and makes the following key contributions to automated crack detection research:
\begin{itemize}
    \item Large Scale Data Consolidation: We integrate multiple public datasets into a single, harmonized resource covering diverse surface types, structural contexts, and geographic regions.
    \item Standardized Multiclass Taxonomy: Heterogeneous labels from all source datasets are harmonized into a nine class taxonomy, which enables meaningful assessment and dataset comparisons.
    \item Standardized Benchmark: To establish baseline classification benchmarks using contemporary DL architectures (DenseNet, ResNet, EfficientNet, ViTB, VGG) to validate dataset quality and provide reference performance metrics for future comparative studies.
    \item High Quality Preprocessing: Low quality, corrupted, and duplicate images are systematically filtered to ensure a clean, reliable, and high quality dataset.
    \item Open Access and Licensing: Novel dataset is freely available under permissive licenses (CC BY 4.0), allowing researchers worldwide to access, use, and build upon it.
\end{itemize}

\section{Analysis of existing crack damage datasets}
Several publicly available datasets have been developed in recent years to aid research into automated crack damage detection in civil infrastructure. These datasets have contributed significantly to the advancement of CV based inspection systems by allowing for supervised learning and performance evaluation. However, a closer look at available datasets indicates significant limits in terms of size, diversity, annotation consistency, and real world representativeness. 

Table \ref{tab:existing_datasets} presents a comparative review of the public datasets in the structural defect literature, highlighting their surface type, resolution and scale. The comparison reveals several patterns. Pavement damage datasets, notably the RDD series, dominate the landscape in terms of scale, with RDD2022 \cite{Rdd2022} alone containing over 47,000 images. However, these datasets are entirely road focused and use a detection oriented annotation approach (bounding boxes with road damage type codes) that does not transfer well to classification tasks or extend to non pavement situations. Concrete focused datasets like METU \cite{METU} and SDNET2018 \cite{Sdnet2018} provide large scale binary classification data, but they are limited to a particular material type and do not provide multi class damage characterization.

\begin{table}
\caption{Comparison of existing datasets with total samples and unique classes}\label{tab:existing_datasets}%
\begin{tabular}{@{}lllllll@{}}
\toprule
Author & Year & Country & Dataset Name & Surface & Resolution & Images\\
\midrule
\cite{Historicbuildingdataset} & 2024 & Iran \& Portugal & Crack in Historic Buildings & Masonry materials & 3456 × 3456 & 4374 \\
\cite{METU} & 2018 & Turkey & METU Concrete Crack & Concrete & 227 x 227 & 40,000 \\
\cite{CONCORNET2023} & 2023 & Mexico & CONCORNET2023 & Reinforced concrete & Variable & 790 \\
\cite{Sdnet2018} & 2018 & USA & SDNET2018 & Concrete & 256 x 256 & 56,000 \\
\cite{CrackNJ156} & 2022 & China & CrackNJ156 & Pavement & 512 × 512 & 156 \\
\cite{CrackSC} & 2023 & United States & CrackSC & Pavement & 320 × 480 & 197 \\
\cite{Cracktree} & 2012 & China & CrackTree206 & Pavement & 800 × 600 & 206 \\
\cite{GAPsV1} & 2017 & Germany & GAPs v1 & Asphalt & 1920 × 1080 & 1,969 \\
\cite{GAPsV2} & 2019 & Germany & GAPs v2 & Asphalt & 1920 × 1080 & 2468 \\
\cite{SUT-Crack} & 2023 & Iran & SUT-Crack & Asphalt pavement & 3024 × 4032 & 130 \\
\cite{PillowDamBorehole} & 2024 &  China & Pillow Dam Borehole CCTV & Internal dam concrete & 1048 × 7098 & 192 \\
\cite{MSD-Det} & 2025 & China & MSD-Det Dataset & Masonry structure & Variable & 1082 \\
\cite{Brickwork-Cracks} & 2023 & United Kingdom & Brickwork Cracks Dataset & Masonry & 227 × 227 & 700 \\
\cite{bridge-crack-detection} & 2019 & China & Bridge Crack Detection & Bridge & 224 × 224 & 6,069 \\
\cite{SVRDD} & 2023 & China & SVRDD & Road & 1024 × 1024 & 8000 \\
\cite{concrete-pavement} & 2025 & Costa Rica & Concrete Pavement Damage & Pavement & 2448 x 1832 & 1178 \\
\cite{concrete-pavement}& 2025 & Costa Rica & Concrete Pavement Slab & Pavement & 2448 x 1832 & 688 \\
\cite{RDD4D} & 2025 & Oman & DRDD(Diverse Road Damage) & Road & 1920 × 1440 & 1500 \\
\cite{RDD} & 2024 & Japan & RDD(Road Damage Detection) & Road & 1024 × 1024 & 1385 \\
\cite{Bridge-crack-library} & 2021 & China & Bridge Crack Library & Concrete, steel, masonry & 256 × 256 & 11000 \\
\cite{CrSpEE} & 2021 & Multi country & CrSpEE & Buildings and bridges & Variable & 2229 \\
\cite{PEER-Hub-ImageNet} & 2020 & Global & PEER Hub ImageNet & Concrete and steel & Variable & 36,413 \\
\cite{RDD2018} & 2018 & Japan & RDD2018 & Road & 600 × 600 & 9053 \\
\cite{RDD2019} & 2019 & Japan & RDD2019 & Road & 600 × 600 & 13,133 \\
\cite{Rdd2020} & 2020 & Multi country & RDD2020 & Road & Vatiable & 26,336 \\
\cite{Rdd2022}&  2022 & Multi country & RDD2022 & Road & Variable & 47,420 \\
\cite{dacl10k} & 2024 & Germany & dacl10k Dataset & Concrete, steel, masonry & Variable & 9,920 \\
\cite{S2DS} & 2022 & Germany & S2DS (Structural Defects) & Concrete & 1024 × 1024 & 743 \\
\cite{Concreteandpavements} & 2023 & Nigeria & Concrete \& Pavement Crack & Pavement/Concrete & 227 x 227  & 30,000 \\
\cite{Crackvision12k} & 2024 & Global & CrackVision12K & Various & 256 × 256 & 12,000 \\
\cite{CODEBRIM} & 2019 & Greece & CODEBRIM & Bridge/Concrete & Variable & ~1,590 \\
\cite{CrackIT} & 2016 & Portugal & CrackIT & Pavement & 2048 x 1536 & 84 \\
\botrule
\end{tabular}
\end{table}

Bridge focused datasets including bridge crack \cite{bridge-crack-detection} and CODEBRIM \cite{CODEBRIM} represent important contributions for infrastructure inspection but cover a narrow structural domain. Heritage building datasets are particularly underrepresented in the literature despite the significant inspection needs of historic built assets. Noncrack damages such as spalling, corrosion, efflorescence, and biological development have little coverage in most existing resources, with CONCORNET2023 \cite{CONCORNET2023} and the dacl10k \cite{dacl10k} dataset being among the few that provide meaningful corrosion and spalling annotations.

Image resolution varies considerably across existing datasets, reflecting differences in capture hardware, deployment context, and preprocessing conventions. Road damage datasets, such as the RDD series \cite{RDD2018, RDD2019, Rdd2020, Rdd2022}, were collected using vehicle mounted smartphones, yielding images at approximately 512×512 to 3650×2044 pixels pixels. While METU \cite{METU} and Brickwork Cracks \cite{Brickwork-Cracks} datasets provide images at lower, fixed resolutions of 256×256 pixels, facilitating rapid annotation and model training. Bridge and structure focused datasets, such as dacl10k \cite{dacl10k} and CODEBRIM \cite{CODEBRIM}, typically have greater native resolutions due to the employment of dedicated inspection cameras. This resolution heterogeneity is a practical challenge for multisource training since naive pooling of pictures from diverse sources generates resolution induced distribution shifts, which can degrade model performance.

Overall, the existing datasets show a fragmented landscape in which no single resource provides the scale, class diversity, resolution consistency, and geographic breadth needed to train generalizable structural damage inspection models. StructDamage dataset directly solves these shortcomings by combining 32 diverse sources into a single, harmonized benchmark that covers 9 surface damage categories across a wide range of structural materials, systems, and geographic regions.

\subsection{Class composition and imbalance analysis across sources}
\begin{figure}
    \centering
    \includegraphics[width=0.9\linewidth]{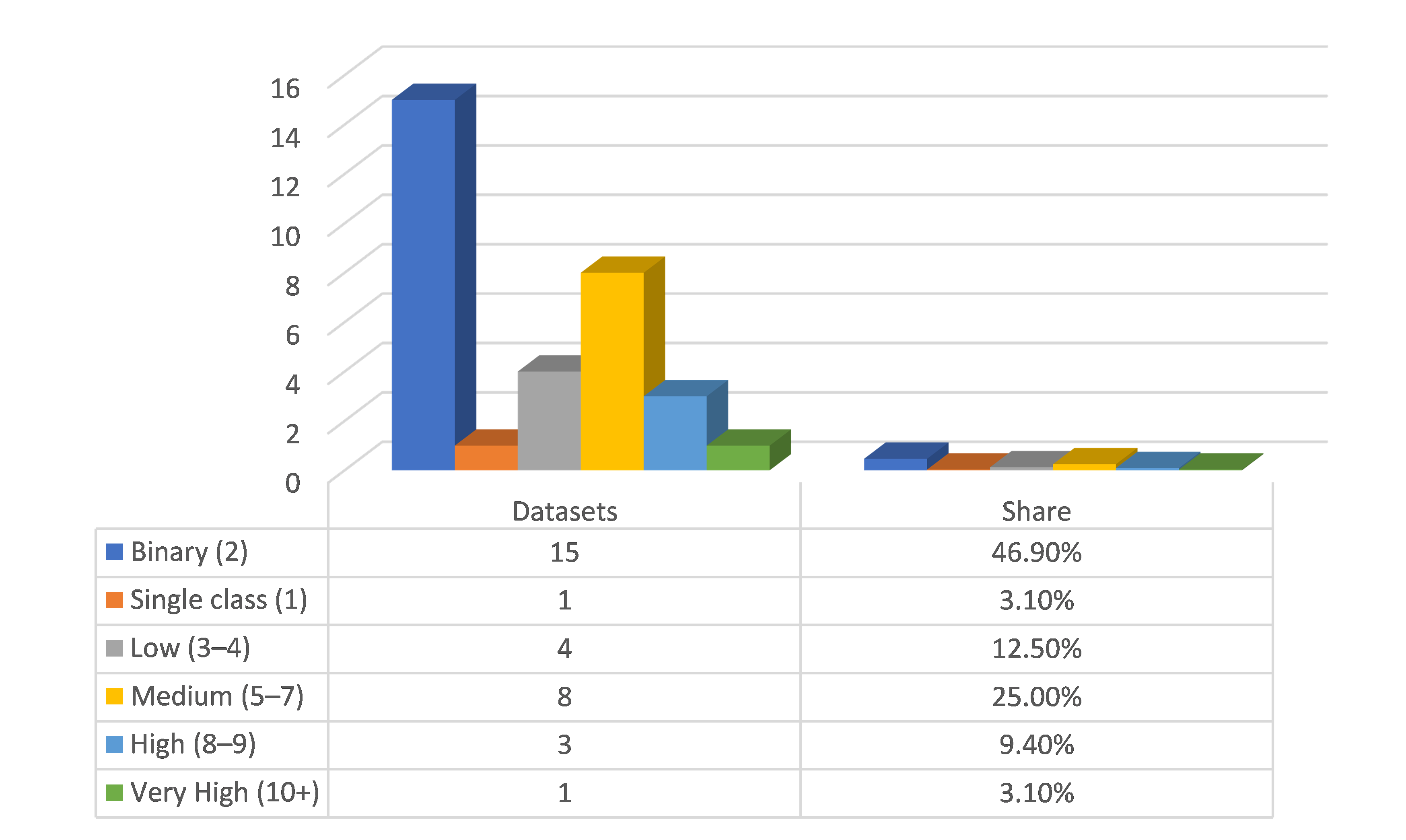}
    \caption{Distribution of calss count categories of source datasets}
    \label{fig: class count}
\end{figure}

Figure \ref{fig: class count} shows the distribution of original class taxonomy sizes among the 32 source datasets used in StructDamage, which are divided into six groups ranging from single class collections to very detailed multiclass schemes. The distribution shows a strong preference for basic taxonomies: 16 of the 32 datasets (50\%) use a binary or single class structure, giving images to only two categories, most typically crack vs nocrack or crack versus background. This preponderance of binary labelling reflects the historical conceptualization of structural damage detection as a detection rather than classification task, with the primary goal of localizing any damage rather than distinguishing between damage kinds.

Four datasets (12.5\%) define 3–4 classes, typically extending the binary scheme with a small number of crack subtypes or material categories (e.g. \cite{Historicbuildingdataset} with Brick, Stone, Cob, and Tile; \cite{Bridge-crack-library} with non steel crack, steel crack, and noise). A further eight datasets (25\%) fall in the medium range of 5–7 classes, including the GAPs series \cite{GAPsV1, GAPsV2} (6 road distress types), \cite{SVRDD} (7 road damage types), \cite{MSD-Det} (7 masonry defect types), and  \cite{S2DS} (7 structural defect categories). These mid range datasets are the most directly transferable to the proposed taxonomy, as their class granularity closely matches the nine (category) scheme and their labels could often be mapped with minimal ambiguity.

Only three datasets (9.40\%) define eight or more classes. \cite{RDD2018} and \cite{PEER-Hub-ImageNet} each define 8 categories, \cite{RDD2019} defines 9, and \cite{dacl10k} defines 19 classes, the most granular taxonomy in the entire collection, encompassing 13 damage types and 6 structural object classes across bridge inspection imagery. This distribution highlights a fundamental mismatch between the binary framing prevailing in the prior literature and StructDamage nine class classification target. The majority of source datasets contribute to only one or two classes, regardless of their original taxonomy size. This structural limitation, rather than annotation granularity, is the major cause of the class coverage asymmetry,the central motivation for the multisource aggregation strategy at the core of StructDamage. The resulting class distribution in novel dataset, while not perfectly balanced, provides significantly more uniform coverage than any single current resource, as demonstrated in section \ref{sec4.2}.

\section{Unified dataset construction pipeline} \label{sec3}

The construction of novel StructDamage dataset from 32 heterogeneous source datasets required a systematic harmonization methodology addressing four distinct challenges: (1) principled selection of source datasets from the broader literature, (2) reconciliation of incompatible annotation schemes into a unified label taxonomy, (3) navigation of diverse licensing terms to ensure legitimate redistribution, and (4) standardized preprocessing to ensure consistent image quality across sources. 
\begin{figure}[h!]
    \centering
    \includegraphics[width=0.8\linewidth]{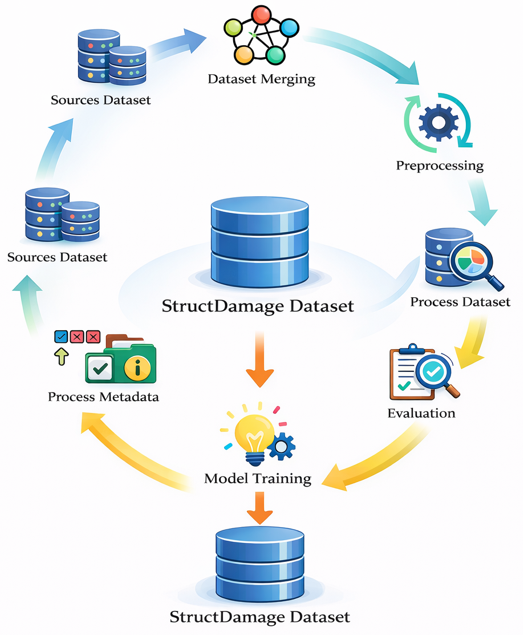}
    \caption{StructDamage dataset construction workflow}
    \label{fig:workflow}
\end{figure}

Figure \ref{fig:workflow} depicts the full workflow used to create the unified StructDamage dataset, illustrating the sequential steps from multisource data gathering and integration to preprocessing, metadata refinement, model driven validation, and final dataset compilation. Each of these components is described in detail in the following subsections.

\subsection{Dataset acquisition \& selection criteria}

The proposed unified novel StructDamage dataset was created from publically available datasets that are commonly utilized in structural damage detection studies. The fundamental goal of the selection procedure was to create a broad coverage across damage categories, structural materials, geographic regions, and image acquisition conditions while maintaining legal compliance, annotation reliability, and practical usefulness for learning based systems.

\begin{figure}[h!]
    \centering
    \includegraphics[width=0.9\linewidth]{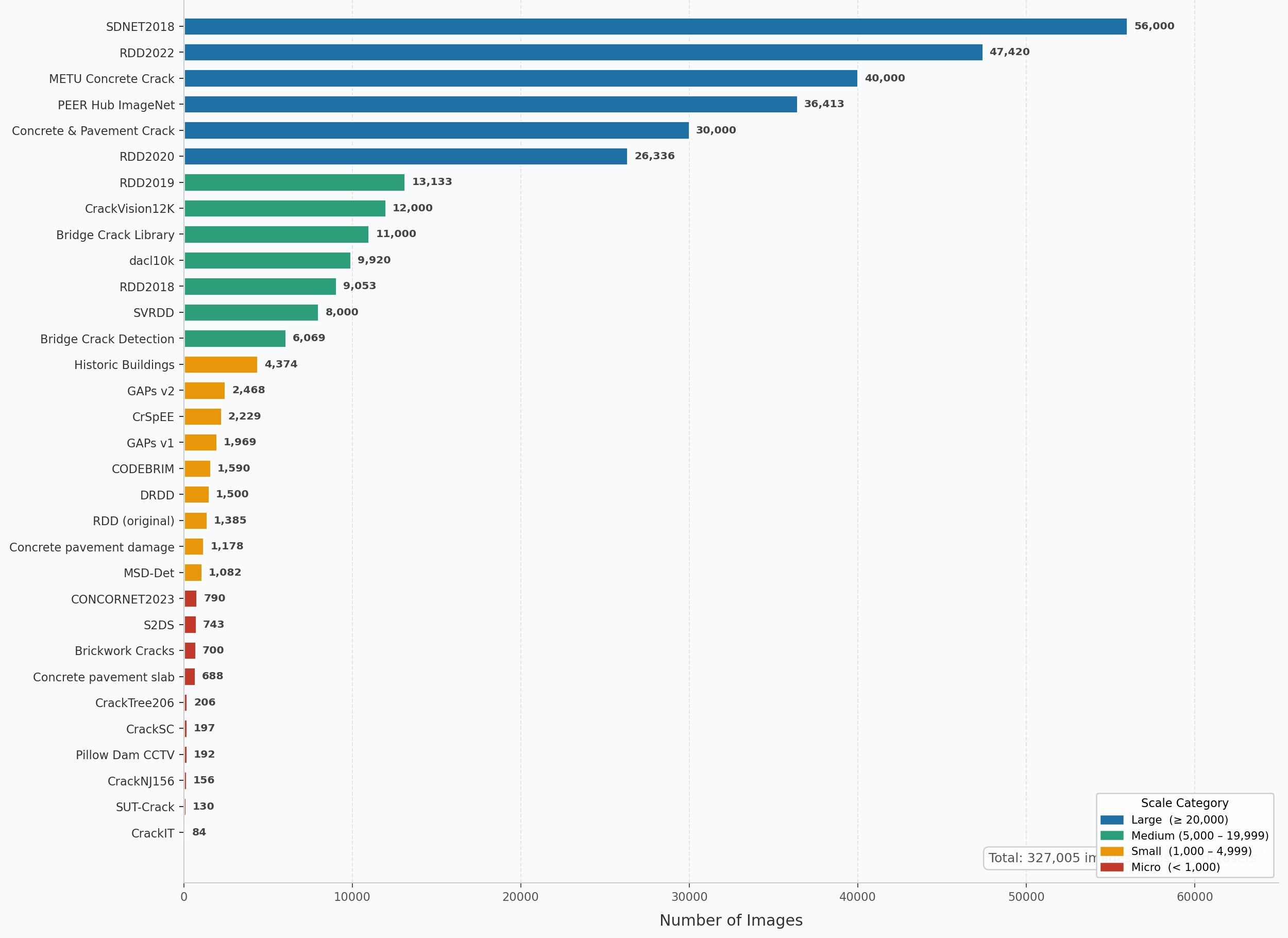}
    \caption{Number of images per source dataset included in StructDamage dataset}
    \label{fig:dataset size}
\end{figure}

The distribution of dataset sizes among selected sources was examined to determine each dataset's contribution to the unified collection. Figure \ref{fig:dataset size} shows that the source datasets vary widely in size, ranging from small, high resolution collections with detailed annotations to large databases comprising tens of thousands of images. Rather than emphasizing size alone, this work takes a balanced approach, with large datasets providing statistical robustness and smaller datasets contributing structural and visual variation. This distributional analysis helps prevent a single dataset from dominating the unified collection and reduces sampling bias.

\begin{figure}[h!]
    \centering
    \includegraphics[width=0.8\linewidth]{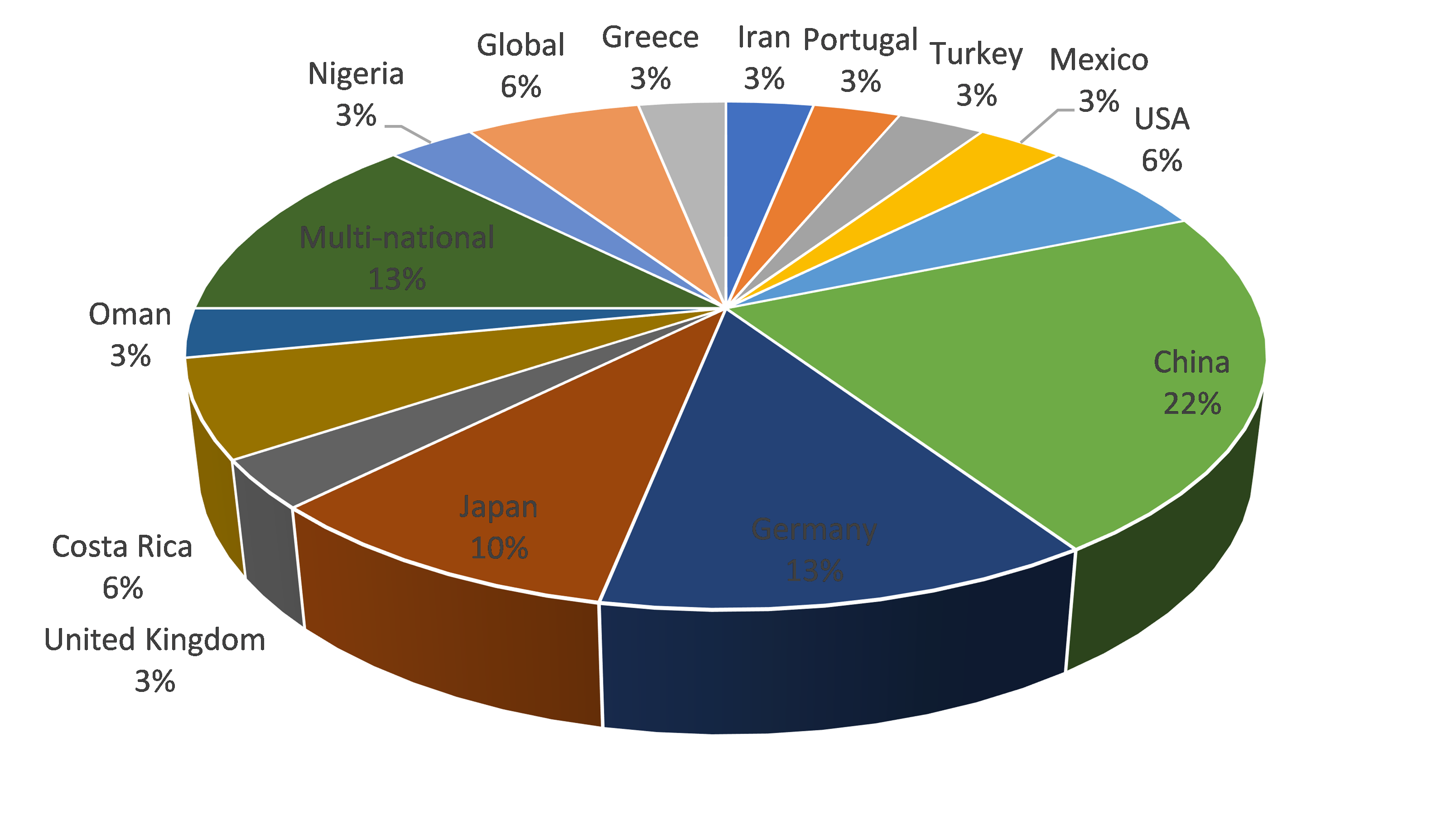}
    \caption{Geographic distribution of source datasets included in StructDamage dataset}
    \label{fig:geographic regions}
\end{figure}

The geographic distribution of the source datasets was investigated to determine regional diversity. Figure \ref{fig:geographic regions} depicts the geographic coverage of the selected datasets, which include different countries and regions from Asia, Europe, and North America. Geographic diversity is especially essential for identifying crack damage, as construction materials, structural design practices, environmental exposure, and aging mechanisms vary across locations. Combining records from multiple geographic settings, the unified dataset captures a broader range of crack features and surface conditions, making it more suitable for cross domain and real world applications.

Aside from size and geographic coverage, the data gathering procedures used by the source datasets were thoroughly examined. Table \ref{tab:collection methods} outlines the acquisition platforms and inspection settings employed in the chosen datasets, which include handheld cameras, mobile devices, vehicle mounted systems, and stationary inspection configurations. These datasets were collected under both controlled and uncontrolled conditions, with varying angles, illumination, and inspection distances. Preserving this variation in acquisition conditions allows the unified StructDamage dataset to reflect realistic inspection environments, facilitating the development of robust crack detection models.
\begin{table}[htbp]
\centering
\caption{Licensing framework and redistribution terms for source datasets}
\label{tab:collection methods}
\small
\begin{tabular}{@{}p{3cm} p{4cm} p{5cm}@{}}
\toprule
\textbf{Source datasets} & \textbf{Collection methods} & \textbf{Captures devices} \\ \midrule
\cite{Historicbuildingdataset, METU, CONCORNET2023, Sdnet2018, CrackNJ156, CrackSC, SUT-Crack, MSD-Det, concrete-pavement, Bridge-crack-library, dacl10k, S2DS, Concreteandpavements, CODEBRIM, CrackIT, bridge-crack-detection} & Ground survey & Smartphone camera, DSLR camera, digital camera, area array CCD camera, drones \\
\cite{GAPsV1, GAPsV2, RDD4D, RDD, RDD2018, RDD2019, Cracktree} & Vehicle mounted & CCD cameras, Smartphone (Android) \\
\cite{PillowDamBorehole} & CCTV / borehole camera & CCTV camera \\
\cite{Brickwork-Cracks} & Ground survey + online sources & Smartphone \& Manufacturer websites \\
\cite{SVRDD} & Street view platform & Street-view camera \\
\cite{CrSpEE} & Post event survey & Smartphone / DSLR \\
\cite{Rdd2022} & Crowdsensed & smartphone, vehicle mounted camera, Google Street View images \\
\cite{PEER-Hub-ImageNet, Crackvision12k} & Multi source aggregation \& Crowdsourced & DSLRs, Smartphones, Laboratory CCDs, UAVs
\\ \bottomrule
\end{tabular}
\end{table}

\subsection{Labeling strategy \& expert validation}

\begin{figure}
    \centering
    \includegraphics[width=0.8\linewidth]{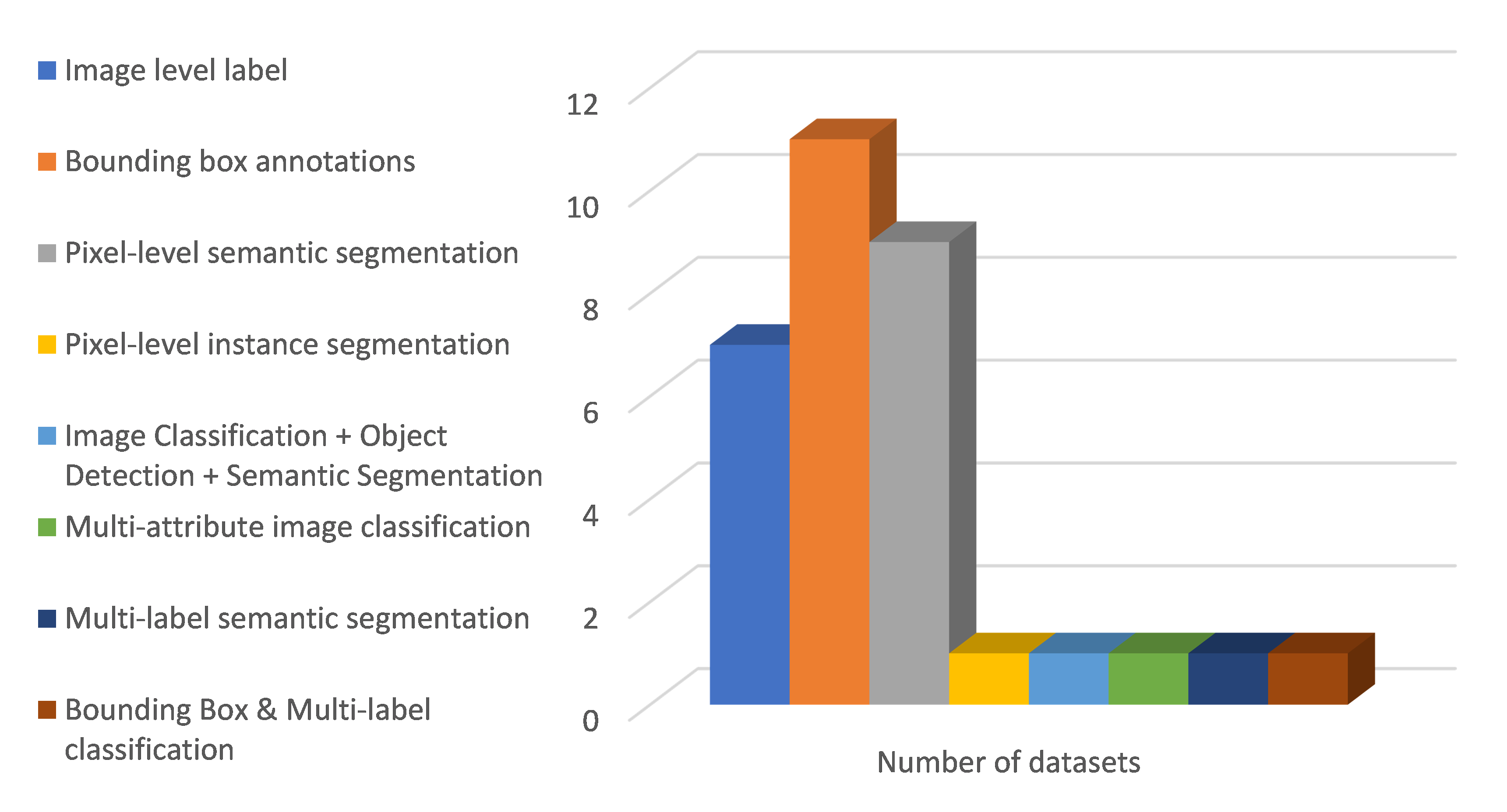}
    \caption{Annotation types across the source datasets}
    \label{fig:annotation type}
\end{figure}

During dataset construction, a consistent annotation harmonization and verification approach was used to ensure consistency and effectiveness. The source datasets used a variety of annotation formats, including binary image level labels (crack/nocrack), multiclass image level labels, instancelevel annotations, pixel level semantic segmentation masks, and combinations of these as represented in table \ref{tab:annotation type}. To better demonstrate the variability of labeling systems across the combined source datasets, we present a graphic comparison of the various annotation formats initially used in Figure \ref{fig:annotation type}.

\begin{table}[htbp]
\centering
\caption{Summary of annotation format used across the source datasets}
\label{tab:annotation type}
\small
\begin{tabular}{@{}p{4.5cm} p{6cm} p{2cm}@{}}
\toprule
\textbf{Source datasets} & \textbf{Annotation type} & \textbf{Number of datasets} \\ \midrule
\cite{Historicbuildingdataset, METU, Sdnet2018, PillowDamBorehole, Brickwork-Cracks, bridge-crack-detection, Concreteandpavements} &  Image level label &  7 \\ 
\cite{CONCORNET2023, GAPsV1, GAPsV2, MSD-Det, SVRDD, RDD4D, RDD, RDD2018, RDD2019, Rdd2020, Rdd2022} &  Bounding box annotations & 11 \\ 
\cite{CrackNJ156, CrackSC, Cracktree, concrete-pavement, Bridge-crack-library, S2DS, Crackvision12k, CrackIT} & Pixel level semantic segmentation & 9 \\ 
\cite{CrSpEE} & Pixel level instance segmentation & 1\\
\cite{SUT-Crack} & Image Classification + Object Detection + Semantic Segmentation & 1 \\
\cite{PEER-Hub-ImageNet} & Multi attribute image classification & 1\\
\cite{dacl10k} & Multi label semantic segmentation &1 \\
\cite{CODEBRIM} & Bounding Box \& Multi label classification & 1
\\ \bottomrule
\end{tabular}
\end{table}

Images containing image level annotations were labeled based on the presence or absence of visible cracks. Detection based datasets with bounding boxes were examined to ensure spatial alignment and compatibility with the unified label definition. Pixel level segmentation masks were evaluated to ensure that fracture regions were accurately identified, particularly for thin or discontinuous cracks. To improve annotation reliability, a verification step was performed on a subset of each source dataset. This approach included visual inspection to detect labeling discrepancies, confusing crack patterns, and annotation errors introduced during data conversion. Samples with unclear crack visibility, serious annotation mistakes, or incompatible labeling were fixed when possible or removed from the final dataset.

Overall, the annotation technique focuses on consistency, task flexibility, and quality assurance. StructDamage provides a dependable and extensible labeling base for crack damage detection research by harmonizing heterogeneous annotations while maintaining their original semantic purpose.
\subsection{Compliance and ethical distribution}

The StructDamage dataset was created in a rigorous commitment to the legal and license restrictions of the original source datasets. Given that the unified dataset contains data from different publicly available repositories, license compatibility, usage rights, and attribution requirements were carefully considered throughout the dataset building process.

Each source dataset was manually checked to determine its licensing terms, including rights for academic use, redistribution, and derivative works. Only datasets with clearly stated licenses for research use were included in the consolidated dataset. The licensing constraints for each source dataset were obtained from their official repository pages, associated publications, and, where necessary, direct interaction with the dataset authors. Table \ref{tab:dataset_licenses} summarizes the license types, usage permissions, and redistribution conditions associated with each dataset.

\begin{table}[htbp]
\centering
\caption{Licensing framework and redistribution terms for source datasets}
\label{tab:dataset_licenses}
\small
\begin{tabular}{@{}p{2.5cm} p{7cm} p{2.5cm}@{}}
\toprule
\textbf{License type} & \textbf{Redistribution terms} & \textbf{Source datasets} \\ \midrule
CC BY 4.0 & Attribution required; commercial and derivative use permitted & \cite{METU, Sdnet2018, CrackNJ156, SUT-Crack, Brickwork-Cracks, SVRDD, RDD, Rdd2022} \\ \addlinespace
CC BY-NC 3.0 & Attribution required; non-commercial use only & \cite{Rdd2020, dacl10k} \\ \addlinespace
CC BY-SA 4.0 & Attribution required; commercial and derivative use permitted; redistribution allowed under the same license & \cite{Concreteandpavements} \\ \addlinespace
CC BY-NC-SA 4.0 & Attribution required; non commercial use only; derivative works permitted; redistribution under the same license (ShareAlike)& \cite{concrete-pavement, PEER-Hub-ImageNet} \\ \addlinespace
MIT license & Attribution required (license notice); commercial use, modification, and redistribution permitted with no copyleft requirement. & \cite{CONCORNET2023, CrackSC, CrSpEE, RDD2018, RDD2019}\\ \addlinespace
Apache-2.0 license & Attribution required (license and NOTICE file); commercial use, modification, and redistribution permitted; patent rights granted. & \cite{RDD4D} \\ \addlinespace
GPL-3.0 license & Attribution required; commercial and derivative use permitted; redistribution allowed under the same GPL-3.0 license (copyleft) & \cite{S2DS} \\ \addlinespace
Research / Academic Only & Redistribution permitted for non commercial research with citation & \cite{CODEBRIM, CrackIT} \\ \addlinespace
CC0 1.0 & No attribution required; commercial and derivative use permitted; redistribution fully allowed without restrictions. & \cite{Bridge-crack-library, Crackvision12k} \\ \bottomrule
\end{tabular}
\end{table}

StructDamage dataset as a whole is licensed under the Creative Commons Attribution 4.0 International (CC BY 4.0). Access to the dataset is granted under explicitly defined usage conditions that specify allowed use cases, citation requirements, and redistribution restrictions. Users must credit both this data descriptor and the source datasets for any images derived from those sources. This approach ensures that the research community can utilize the unified dataset securely and responsibly by adhering to established legal and ethical norms and offering transparent access policies. 

\subsection{Data refinement \& formatting}

\begin{figure} [htbp]
    \centering
    \includegraphics[width=0.8\linewidth]{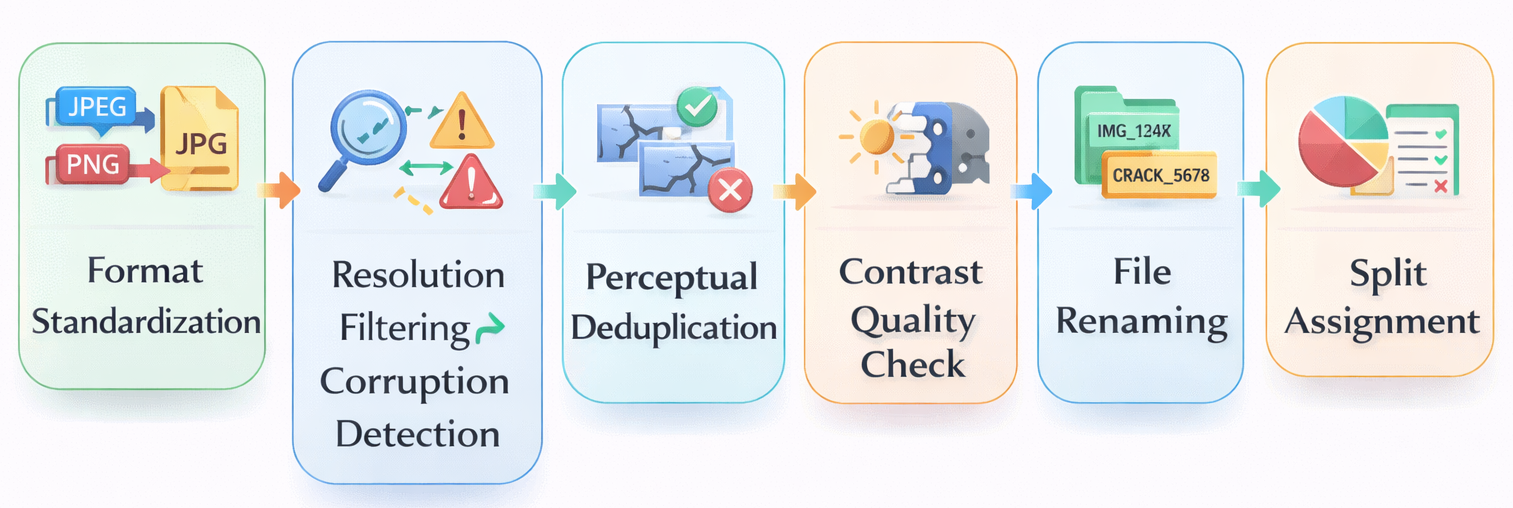}
    \caption{Preprocessing operations}
    \label{fig:preprocessing}
\end{figure}

Prior to final dataset compilation, a systematic preparation procedure was used to assure consistency, quality, and usefulness across diverse source datasets. The preprocessing methods shown in Figure \ref{fig:preprocessing} were intended to standardize image formats, remove noise and redundancy, and prepare the data for reliable downstream use.

A structured preprocessing pipeline was implemented to ensure consistency, quality, and usability across multiple heterogeneous datasets before final integration. First, all images were converted into a unified format to maintain compatibility and simplify processing. Low resolution, corrupted, or unreadable files were then removed to maintain minimum quality standards and prevent training instability.

To eliminate redundancy from overlapping sources, perceptual similarity analysis was used to detect and remove duplicate or near duplicate images, reducing dataset bias. A contrast quality check further filtered out images with poor visibility, excessive noise, or inadequate illumination to preserve clear crack patterns.

Next, annotation formats from different datasets were standardized into a unified schema while retaining their original meaning. Files were renamed using a consistent convention to ensure traceability between images and labels. Finally, the dataset was split into training, validation, and test sets in a balanced manner to prevent data leakage and support fair evaluation. Overall, the pipeline produces a clean, standardized, and reliable dataset suitable for crack detection research.

\section{Dataset attributes}
StructDamage dataset consists of approximately 78093 images collected from 32 publicly available datasets. The images represent diverse structural materials captured under varying lighting, environmental, and geographic conditions. Each image is associated with a  standardizedlabel based on the proposed nine class damage taxonomy. Labels are provided in a consistent image level format to ensure interoperability across tasks. The corresponding metadata files include class labels and source dataset information. All annotations, preparation scripts, and benchmark training code are given to ensure consistency and ease of usage.
\subsection{Directory Architecture}
Novel StructDamage dataset is organized in a simple, standard format for classification tasks. The dataset is split into training, validation, and test splits in an 80:10:10 ratio. Each split has subdirectories for the nine classes described in the unified taxonomy. This hierarchical structure enables ease of use, reproducibility, and smooth connection with common DL frameworks.
\subsection{Class distribution and semantic definitions} \label{sec4.2}
\begin{figure}
    \centering
    \includegraphics[width=0.9\linewidth]{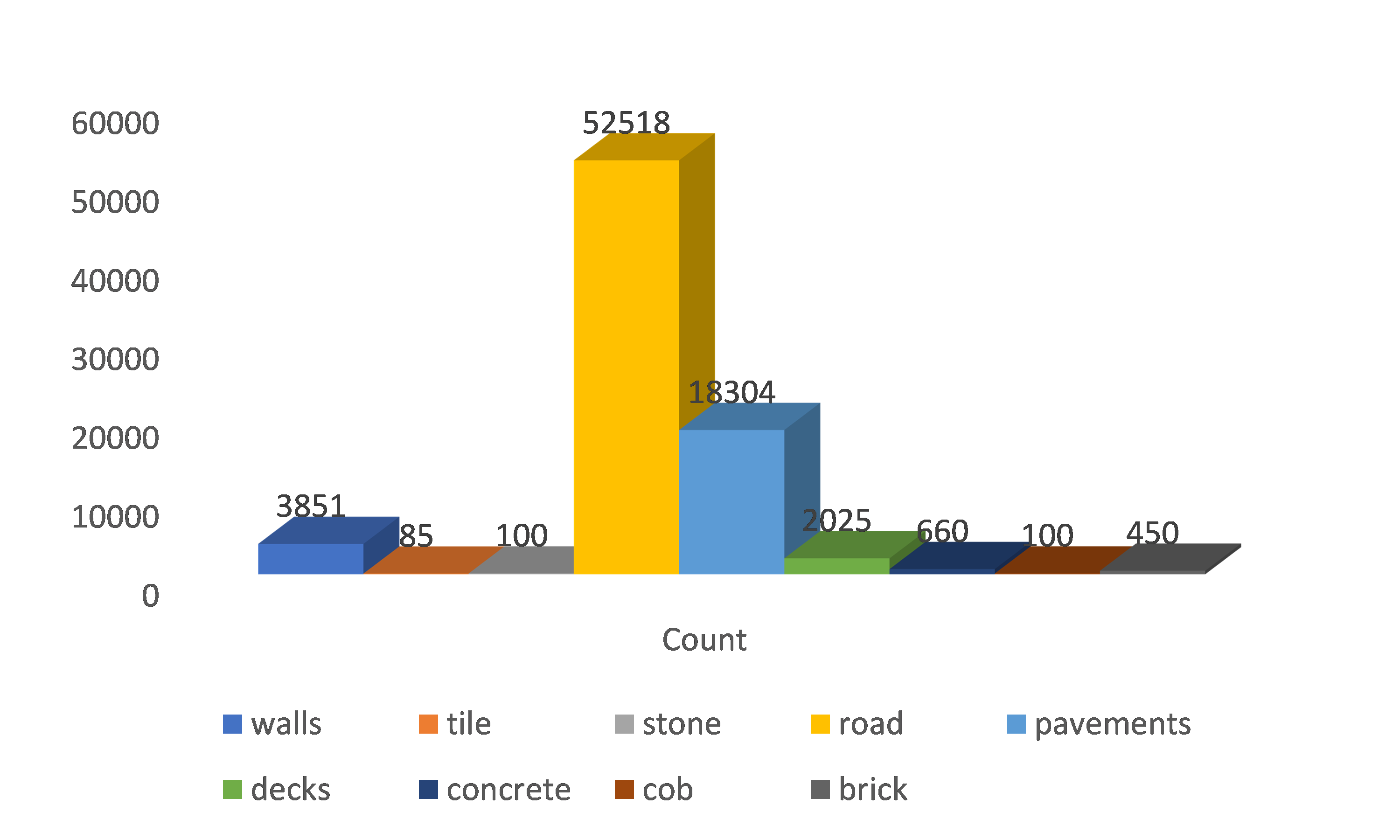}
    \caption{Class distribution of StructDamage dataset}
    \label{fig:class categories}
\end{figure}

Unified StructDamage dataset includes precise class level data as well as clear semantic descriptions for each form of damage to ensure balanced representation and relevant benchmarking. The dataset is classified into nine categories: walls, tile, stone, road, pavements, decks, concrete, and brick damage categories. Figure \ref{fig:class categories} illustrates the distribution of images per class, highlighting the relative frequency of each surface damage type. While road cracks constitute the largest category, the dataset includes sufficient samples of underrepresented damage types such as tile and deck to support robust model generalization.

Each class is defined based on established infrastructure inspection standards and visual characteristics. 

\section{Technical validation}
This section discusses the technical validation trials used to establish a baseline classification performance on the StructDamage dataset. The validation consists of a controlled preprocessing technique aimed to solve the significant class imbalance found in the raw aggregated dataset, followed by a systematic evaluation of sixteen modern DL architectures covering five model families. The goal of these tests is to prove that StructDamage has enough discriminative information to permit high accuracy structural damage classification and to offer reliable reference standards against which future approaches can be evaluated.
\subsection{Preprocessing for baseline experiments} \label{sec5.1}
Before training, the raw StructDamage dataset requires further balancing beyond the harmonization described in \ref{sec3}, because the assembled dataset's class distribution reflects the availability of source imagery rather than a purposeful sampling scheme. Road damage class, in particular, contains more than 50,000 images for longitudinal and transverse cracks combined, while the concrete pavement class contributes more than 18,000 images, resulting in a severe interclass imbalance that would cause models to overfit to the majority of classes during training. To overcome this imbalance, two complimentary procedures were implemented: similarity based subsampling for majority classes and data augmentation for minority classes.
\subsubsection{Similarity based subsampling of majority classes}
To eliminate redundancy while maintaining visual diversity, a perceptual similarity filter was applied to classes with sample counts exceeding the target size. Each image in an overrepresented class was assigned a perceptual hash (pHash), and pairwise Hamming distances were calculated against all previously retained images. An image was preserved if its highest similarity to any previously retained image was less than 30\%. Images with less than 30\% similarity index to all retained samples were kept, ensuring that only visually unique instances were included. This threshold was empirically selected to exclude near duplicate and highly repetitive frames in vehicle mounted survey film while retaining morphologically distinct crack appearances across a range of surface textures, lighting conditions, and severity levels.
\subsubsection{Augmentation of minority classes}
Classes with sample counts less than the target threshold of 5,000 images were enhanced with a combination of geometric and photometric adjustments to increase training diversity. Geometric augmentations included random horizontal and vertical flips, rotations ranging from -70° to +70°, random cropping with resizing, and perspective distortion. Photometric enhancements included random brightness and contrast jitter, Gaussian blur, and hue saturation perturbation. To minimize storing redundant copies and maximize stochastic variation between epochs, all augmentations were performed online during training rather than offline preprocessing. Augmented samples were generated until each minority class reached a target of around 5,000 images, ensuring that each class had a comparable number of separate training instances.
\subsubsection{Resulting balanced distribution and data splits}
\begin{figure}
    \centering
    \includegraphics[width=0.9\linewidth]{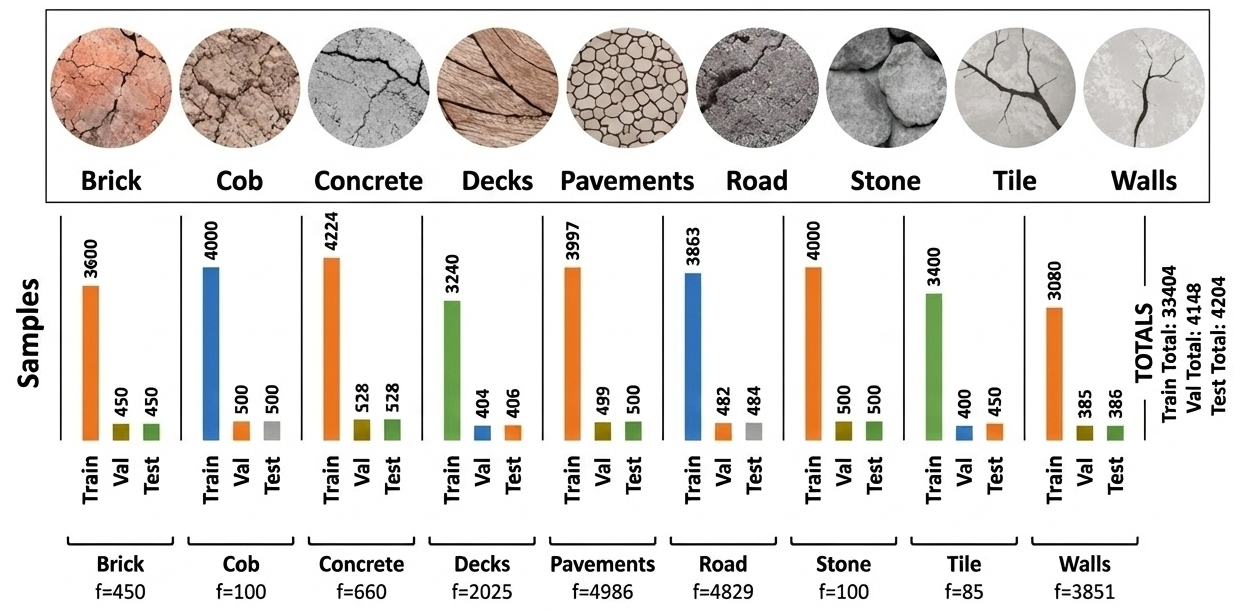}
    \caption{Balanced data distribution}
    \label{fig: balanced data}
\end{figure}

After applying similarity based subsampling to majority classes and augmentation to minority classes, each of the nine damage categories has around 5,000 images, resulting in a balanced corpus of around 41,756 images for the validation trials. This balanced dataset was divided into training, validation, and test splits with ratios of 80\%, 10\%, and 10\%, respectively, and stratified at the image level to maintain class proportions throughout all three subsets. The resulting splits include around 33,404 training images, 4,148 validation images, and 4,204 test images as represented in Figure \ref{fig: balanced data}. It should be noted that this balanced subset is only used for the baseline validation experiments described in this section; the full StructDamage dataset (78,093 images with original distribution) is the primary release and is recommended for use in downstream research, where investigators can use their own sampling and augmentation strategies.

\subsection{Baseline results}
Table \ref{baseline results} shows the classification performance of fifteen architectures across six model families: DenseNet, EfficientNetV2, MobileNetV2, ResNet V2, VGG, and Vision Transformers (ViT). The results are shown as accuracy (\%), precision, recall, and F1-score. The best results for each statistic are displayed in bold.

\begin{table}[]
\caption{Baseline classification results on StructDamage (balanced validation subset)}
\label{baseline results}
\begin{tabular}{lllll}
\toprule
Methods          & Accuracy (\%)    & Precision (\%)   & Recall (\%)      & F1-Score (\%)    \\
\midrule
DenseNet121      & 98.4300          & 0.98326          & 0.98378          & 0.98339          \\
DenseNet169      & 98.4300          & 0.98286          & 0.98334          & 0.98302          \\
DenseNet201      & \textbf{98.6203} & \textbf{0.98526} & \textbf{0.98580} & \textbf{0.98534} \\
EfficientNetV2B0 & 96.2654          & 0.96060          & 0.96047          & 0.96000          \\
EfficientNetV2B1 & 93.9105          & 0.93746          & 0.93820          & 0.93670          \\
MobileNetV2      & 98.3349          & 0.98220          & 0.98236          & 0.98214          \\
ResNet50V2       & 98.3111          & 0.98192          & 0.98251          & 0.98207          \\
ResNet101V2      & 98.3349          & 0.98218          & 0.98228          & 0.98219          \\
ResNet152V2      & 98.2397          & 0.98148          & 0.98139          & 0.98128          \\
VGG16            & 94.8858          & 0.94697          & 0.94651          & 0.94568          \\
VGG19            & 94.8382          & 0.94618          & 0.94624          & 0.94509          \\
VITB16           & 98.0494          & 0.97902          & 0.97966          & 0.97915          \\
VITB32           & 98.4300          & 0.98316          & 0.98328          & 0.98319          \\
VITL16           & 98.3306          & 0.98215          & 0.98210          & 0.98222          \\
VITL32           & 96.7174          & 0.96484          & 0.96532          & 0.96480         \\
\bottomrule
\end{tabular}
\end{table}

The results show that StructDamage can perform high accuracy multiclass structural damage classification over a wide range of architectures. Twelve of the fifteen assessed models obtain F1-scores greater than 0.96, indicating that the dataset has enough visual discriminability to permit strong classification using traditional transfer learning procedures.
Among the examined architectures, DenseNet201 achieves the highest precision (0.98526), recall (0.98580), and F1-score (0.98534), making it the best performing model overall. DenseNet121, DenseNet169 and ViT-B/32 follow closely, with F1-scores of 0.98339, 0.98302 and 0.98319, respectively. Within the DenseNet family, performance continuously increases with depth, with DenseNet201 surpassing DenseNet169 and DenseNet121. The ResNet V2 family performs consistently across all three depth variants (50, 101, and 152 layers), with F1-scores firmly clustered between 0.9812 and 0.9822, indicating that adding depth does not result in significant benefits for this task.

Among Vision Transformer variations, ViT-L/16 and ViT-B/32 attain competitive performance above 0.982 F1-score, whereas ViT-L/32 and ViT-B/16 perform comparably worse at 0.9648 and 0.9792, respectively. This shows that patch resolution has a greater impact than model size on crack morphology recognition, where fine grained spatial detail is crucial for differentiating crack types. The VGG family produces the lowest results of any evaluated family, with VGG16 and VGG19 achieving F1-scores of 0.9457 and 0.9451, respectively, most likely reflecting the absence of residual connections, batch normalization, and modern regularization techniques found in more recently developed architectures. Similarly, EfficientNetV2B1 has the lowest overall F1-score (0.9367), indicating that this version is sensitive to the specific input resolution and capacity configuration used.

Precision and recall are carefully balanced across all architectures, with no model showing a consistent bias toward false positives or false negatives. This balance is comparable with the controlled class distribution used during validation preprocessing, indicating that the similarity-based subsampling and augmentation approach effectively mitigated the class imbalance in the raw dataset. The general distribution of findings, with most designs earning F1-scores more than 0.98, supports StructDamage quality as a benchmark dataset and provides a solid reference baseline for future model development and comparison.
\subsection{Per class evaluation}
To complement the aggregate metrics presented in Table \ref{baseline results}, a comprehensive per class classification report was prepared for the best performing architecture, DenseNet201, tested on the balanced test split (4,204 images). Table \ref{tab:perclass_report} includes per class precision, recall, F1-score, and support, as well as macro and weighted averaged summaries and the Cohen's kappa coefficient. The classification report demonstrates a highly consistent performance profile across all nine damage categories, with the majority of classes obtaining nearly flawless discrimination under controlled balanced conditions.
\begin{table}[ht]
\centering
\caption{Per class classification report}
\label{tab:perclass_report}
\begin{tabular}{llllll}
\hline
\textbf{ID}      & \textbf{Class}     & \textbf{Precision} & \textbf{Recall} & \textbf{F1-Score} & \textbf{Support} \\ \hline
C-00             & Brick              & 1.0000             & 0.9778          & 0.9888            & 450              \\
C-01             & Cob                & 1.0000             & 1.0000          & 1.0000            & 500              \\
C-02             & Concrete           & 1.0000             & 0.9981          & 0.9991            & 528              \\
C-03             & Decks              & 0.9824             & 0.9631          & 0.9726            & 406              \\
C-04             & Pavements          & 0.9876             & 0.9540          & 0.9705            & 500              \\
C-05             & Road               & 0.9958             & 0.9897          & 0.9927            & 484              \\
C-06             & Stone              & 0.9766             & 1.0000          & 0.9881            & 500              \\
C-07             & Tile               & 0.9978             & 1.0000          & 0.9989            & 450              \\
C-08             & Walls              & 0.9272             & 0.9896          & 0.9574            & 386              \\ \hline
\multicolumn{2}{l}{\textbf{Accuracy}} &                    &                 & \textbf{0.9862}   & \textbf{4204}    \\
\multicolumn{2}{l}{\textbf{Macro avg}} & \textbf{0.9853} & \textbf{0.9858} & \textbf{0.9853}   & \textbf{4204}    \\
\multicolumn{2}{l}{\textbf{Weighted avg}} & \textbf{0.9866} & \textbf{0.9862} & \textbf{0.9863}   & \textbf{4204}    \\
\bottomrule
\end{tabular}
\end{table}

The overall test accuracy of 98.62\% and macro-averaged F1-score of 0.9853 are supported by a weighted F1-score of 0.9863, indicating that performance is consistent across classes with different support sizes. The Cohen's kappa coefficient of 0.9845 implies near perfect interclass agreement beyond chance, providing strong evidence that the model developed true discriminative representations for each damage category rather than relying on class frequency bias. The nearly identicality of macro and weighted averages (0.9853 vs. 0.9863) demonstrates that the balanced sample strategy effectively reduced the distributional bias that would otherwise result in exaggerated weighted measures.

\begin{figure}[ht]
    \centering
    \includegraphics[width=0.8\linewidth]{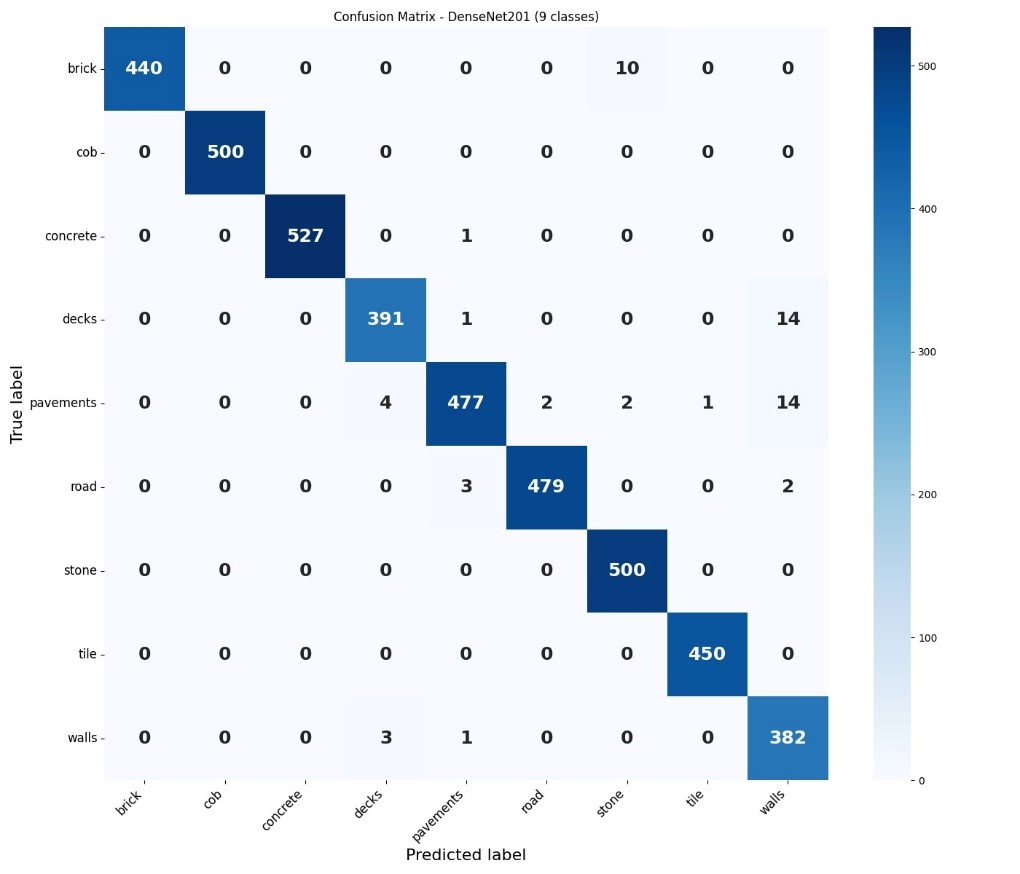}
    \caption{Confusion matrix}
    \label{fig: confusion matrix}
\end{figure}

Figure \ref{fig: confusion matrix} presents the normalized confusion matrix for DenseNet201 evaluated on the balanced StructDamage test split. Each cell (\textit{i,j}) represents the proportion of test samples belonging to true class \textit{i} that were predicted as class \textit{j}, with diagonal entries indicating correct classification rates and off diagonal entries indicating inter class confusion. The matrix provides a granular decomposition of the aggregate metrics reported in Table \ref{tab:perclass_report}, revealing the specific pairwise confusion patterns that underlie the per class precision and recall scores.

The balanced and raw dataset evaluations, taken together, provide complementary viewpoints on StructDamage's value. The balanced validation shows that, under equitable sampling conditions, the dataset is sufficiently discriminative to support near perfect 9 class structural damage classification with standard ImageNet pretrained architectures, which validates the annotation harmonization quality, the preprocessing pipeline effectiveness, and the aggregated source collection's representational richness. In contrast, the raw dataset follows the natural distribution of structural damage pictures in the available literature, with road crack categories being significantly overrepresented in comparison to other classes. Researchers using the full StructDamage release should, therefore, use appropriate class balancing strategies such as class weighted loss functions, oversampling, or the similarity based subsampling protocol described in \ref{sec5.1}, unless their specific research objective requires training under natural distributions. StructDamage is distributed with both the original class distribution and full source metadata to enable both use cases, and the preprocessing code used for the baseline tests is included with the dataset to assure complete replication of the validation results described here.
\section{Usage Notes}
StructDamage is intended for research purposes in the fields of computer vision, structural health monitoring, civil engineering, and related disciplines. Users are encouraged to employ the dataset for training, validation, and benchmarking of machine learning models for structural damage classification. The dataset may also serve as a pretraining corpus for transfer learning to domain specific inspection tasks.

Users should be aware of the following important considerations. First, image quality and acquisition conditions vary significantly across source datasets, reflecting the heterogeneous imaging environments of real world inspections. This variability is a feature rather than a limitation, as it promotes model generalization. Second, class boundaries in real world damage scenarios are not always discrete; some structures may exhibit multiple cooccurring damage types, and the present dataset contains single label annotations for simplicity. Multi label extensions are left for future work. Third, while deduplication was performed at the perceptual hash level, subtle domain shifts between source datasets may influence model generalization; users conducting cross dataset evaluation should account for potential source-level correlations.
\backmatter

\section*{Declarations}
\begin{itemize}
\item \textbf{Ethics approval and consent to participate.}
\item \textbf{Funding.} No funding
\item \textbf{Declaration of competing interest.} The authors declare that they have no known competing financial interests or personal relationships that could have appeared to influence the work reported in this paper.
\item \textbf{Consent for publication.} Not applicable
\item \textbf{Data availability.} Data is publicly available at: \href{https://cloud.dfki.de/owncloud/index.php/s/WNiPcgMnZL9p9rR}{Dataset}
\item \textbf{CRediT authorship contribution statement.} Misbah Ijaz \& Saif Ur Rehman Khan: Conceptualization, Data curation, Methodology, Software, Validation, Writing original draft \& Formal analysis. Muhammed Nabeel Asim, Sebastian Vollmer \& Andreas Dengel: Conceptualization, Funding acquisition, Review. Abd Ur Rehman: Supervision, review \& editing.
\end{itemize}

\bibliography{sn-bibliography}

\end{document}